\documentclass{ecai} 

\usepackage{latexsym}
\usepackage{amssymb}
\usepackage{amsmath}
\usepackage{amsthm}
\usepackage{booktabs}
\usepackage{enumitem}
\usepackage{graphicx}
\usepackage{color}

\newcommand{\BibTeX}{B\kern-.05em{\sc i\kern-.025em b}\kern-.08em\TeX}

\usepackage{ragged2e} 
\usepackage{eso-pic}
\AddToShipoutPicture*{
  \AtTextLowerLeft{
    \put(0,15){%
      \parbox{\textwidth}{%
        \justifying\footnotesize
        This is a preprint of a paper presented at the \textit{European Conference on Artificial Intelligence (ECAI 2025)}. It is made publicly available for the benefit of the research community and should be regarded as a preprint rather than a formally reviewed publication.%
      }%
    }
  }
}

\begin{document}


\begin{frontmatter}


\paperid{5736} 


\title{LLMs as Policy-Agnostic Teammates: A Case Study in Human Proxy Design for Heterogeneous Agent Teams}




\author[A]{\fnms{Aju}~\snm{Ani Justus}\orcid{0009-0002-7125-9397}}
\author[A]{\fnms{Chris}~\snm{Baber}\orcid{0000-0002-1830-2272}} 

\address[A]{School of Computer Science, University of Birmingham}


\begin{abstract}

A critical challenge in modelling Heterogeneous-Agent Teams is training agents to collaborate with teammates whose policies are inaccessible or non-stationary, such as humans. Traditional approaches rely on expensive human-in-the-loop data, which limits scalability. We propose using Large Language Models (LLMs) as policy-agnostic human proxies to generate synthetic data that mimics human decision-making. To evaluate this, we conduct three experiments in a grid-world capture game inspired by Stag Hunt, a game theory paradigm that balances risk and reward. In Experiment 1, we compare decisions from 30 human participants and 2 expert judges with outputs from LLaMA 3.1 and Mixtral 8x22B models. LLMs, prompted with game-state observations and reward structures, align more closely with experts than participants, demonstrating consistency in applying underlying decision criteria. Experiment 2 modifies prompts to induce risk-sensitive strategies (e.g. “be risk averse”). LLM outputs mirror human participants’ variability, shifting between risk-averse and risk-seeking behaviours. Finally, Experiment 3 tests LLMs in a dynamic grid-world where the LLM agents generate movement actions. LLMs produce trajectories resembling human participants’ paths. While LLMs cannot yet fully replicate human adaptability, their prompt-guided diversity offers a scalable foundation for simulating policy-agnostic teammates.

\end{abstract}

\end{frontmatter}


\section{Introduction}

Multi-Agent Reinforcement Learning (MARL) \citep{Albrecht2024} has set the standard for cooperative multi-agent systems, surpassing human performance in games like StarCraft \citep{Vinyals2019} and Go \citep{Silver2017} through self-play and population-based training \citep{Long2023}. Yet, these methods fall short in heterogeneous-agent teams, particularly when humans are involved. While human-AI teams can achieve success in controlled settings like Capture the Flag \citep{Sioutis2004}, this often occurs only after humans unilaterally adapt to AI policies. In our experience with Overcooked-AI, a benchmark environment for fully cooperative human-AI task performance inspired by the popular video game Overcooked \citep{overcookedai2019}, a self-play-trained MARL agent exhibited rigid behaviours, forcing human players to avoid collisions rather than collaborate strategically. This exposes a critical gap, i.e. MARL agents struggle to adapt to policy-agnostic teammates (e.g. humans) with unobservable preferences, strategies, or cognitive constraints.  Large Language Models (LLMs) present a promising approach to synthesise human-like decisions across domains \cite{Argyle2023}, from robotic planning \cite{Huang2022} to reward design \cite{Kwon2023}. However, their reliability as human proxies in heterogeneous-agent reinforcement learning remains unexplored. 


\section{Background}
\label{sec:background}
For heterogeneous teams, key challenges arise from differences in perception, goals, rewards etc. When teammates are human, there is a need to reflect the differences in ability between them and computer agents that have been trained using reinforcement learning.

\textbf{Reinforcement Learning (RL)} enables agents to learn optimal policies through environment interactions. At its core, RL is formalized as a Markov Decision Process (MDP) \citep{sutton2000policy}, defined by the tuple \((S, A, P, R, \gamma)\), where \(S\) is the state space, \(A\) the action space, \(P\) the transition probabilities, \(R\) the reward function, and \(\gamma\) the discount factor. Single-agent RL algorithms, such as Q-learning and policy gradients \citep{schulman2017proximal}, optimize policies \(\pi: S \rightarrow A\) to maximize cumulative rewards.

\textbf{Multi-Agent RL (MARL)} extends RL to settings with interacting agents, modelled as Markov Games \((N, S, \{A_i\}, P, \{R_i\}, \gamma)\). The Centralized Training with Decentralized Execution (CTDE) paradigm \citep{lowe2017multi} addresses inherent non-stationarity by allowing agents to train with access to the global state while constraining them to act only on their local observations at execution time. Algorithms like MADDPG \citep{lowe2017multi} and QMIX \citep{rashid2018qmix} excel in homogeneous teams but struggle with heterogeneous agents that differ in observation spaces, action sets, or reward functions \citep{kuba2024harl}.

\textbf{Heterogeneous-Agent RL (HARL)} frameworks like HAPPO \citep{kuba2024harl} extend policy gradient methods but assume \textit{policy accessibility}, i.e., knowledge of teammate strategies. This assumption breaks down in human-AI teams, where humans exhibit latent preferences (unobservable reward functions \citep{cao2012overview}), cognitive constraints (bounded rationality \citep{Chen2017}), and context-dependent strategies (situational adaptability \citep{overcookedai2019}).

One approach to this challenge involves designing human-like proxies that mimic perceptual, cognitive, and motor constraints. Bounded rationality can be reflected in resource-rational models \cite{Acharya2018, Chen2017, Chen2021} that formalize these constraints, enabling agents to learn responsive policies in simple decision games \cite{Acharya2024}. However, scaling these models to complex tasks (e.g., strategic planning) remains intractable. Alternatives like Reinforcement Learning with Human Feedback (RLHF) \cite{Chaudhari2024} refine policies through iterative human input, but require costly and labor intensive data collection. 

Human-in-the-Loop Reinforcement Learning (HITL RL) is successful in autonomous driving \cite{cao2012overview}, language model fine-tuning \cite{ziegler2019finetuning}, music generation \citep{aju2023music}, and NPC game training \cite{borovikov2019hitlnpc}. However, HITL RL faces data stratification challenges \citep{argall2009survey}.

\textbf{LLMs} offer a scalable alternative to HITL RL by generating training data \citep{Argyle2023}. Recent work demonstrates their ability to simulate consumer choices \citep{Gui2023} and survey responses \citep{Aher2023}, guide RL through reward shaping \citep{Kwon2023} and text-based policy generation \citep{Huang2022}, and act as autonomous agents in text environments \citep{Park2023}.

While prior work has used LLMs to generate synthetic human data, our focus is narrower: we examine whether LLMs can replicate human-like decisions in a well-defined cooperative task without additional training or supervision. Though studies have explored simulating human behaviour with LLMs \citep{Aher2023, Argyle2023, Park2023}, important limitations remain. As \citet{Gui2023} note, LLMs may produce plausible yet incorrect outputs when lacking causal context, a challenge only partially addressed by prompt engineering. Building on this insight, our experimental design tightly constrains LLM outputs to discrete actions and decisions, thereby reducing ambiguity and improving consistency when evaluating alignment with human and expert judgments.

LLMs also face challenges in MARL, including hallucination (plausible but non-human decisions; \citealt{Gui2023}), temporal myopia (poor multi-step planning; \citealt{Park2023}), and risk mismatch (default outputs lacking human risk profiles; \citealt{Aher2023}). 

To reduce hallucination and ensure consistency, we set the temperature parameter to zero, forcing the model to produce deterministic outputs that align with the highest-probability responses. To mitigate temporal myopia, we employ step-by-step prompting at each state of the reinforcement learning environment, encouraging the model to reason through multi-step decision processes. Finally, to account for risk mismatch, we modify prompts to explicitly describe agents with varying risk profiles, and use a high \textit{top-p}, which tells the model to consider only the most likely words up to a probability cutoff $p$, to maintain a diverse range of plausible but human-aligned outputs.

\section{Methodology}
\label{sec:methodology}
We evaluate Large Language Models (LLMs) as policy-agnostic proxies in a \textit{grid-world capture game} inspired by the Stag Hunt paradigm \citep{skyrms2003stag}. This environment provides a controlled testbed for studying human-AI collaboration under partial observability and strategic uncertainty. In the following experiments, we evaluate LLMs as human proxies and policy‑agnostic teammates in a grid‑world stag hunt game, addressing three questions:

\textbf{(Q1) Alignment:} Can LLMs replicate expert decisions with full observability of the environment?

\textbf{(Q2) Adaptability:} Can prompts to LLMs induce human-like response variability and risk sensitivity?

\textbf{(Q3) Human-Proxy Decision Making:} Can LLM agents simulate human-like decision-making and generate coherent multi-step action sequences in a multi-agent team?

\subsection{Grid-World Stag Hunt Environment}
\label{subsec:grid_world}
\begin{figure}[htbp]
    \centering
    \includegraphics[width=0.9\columnwidth]{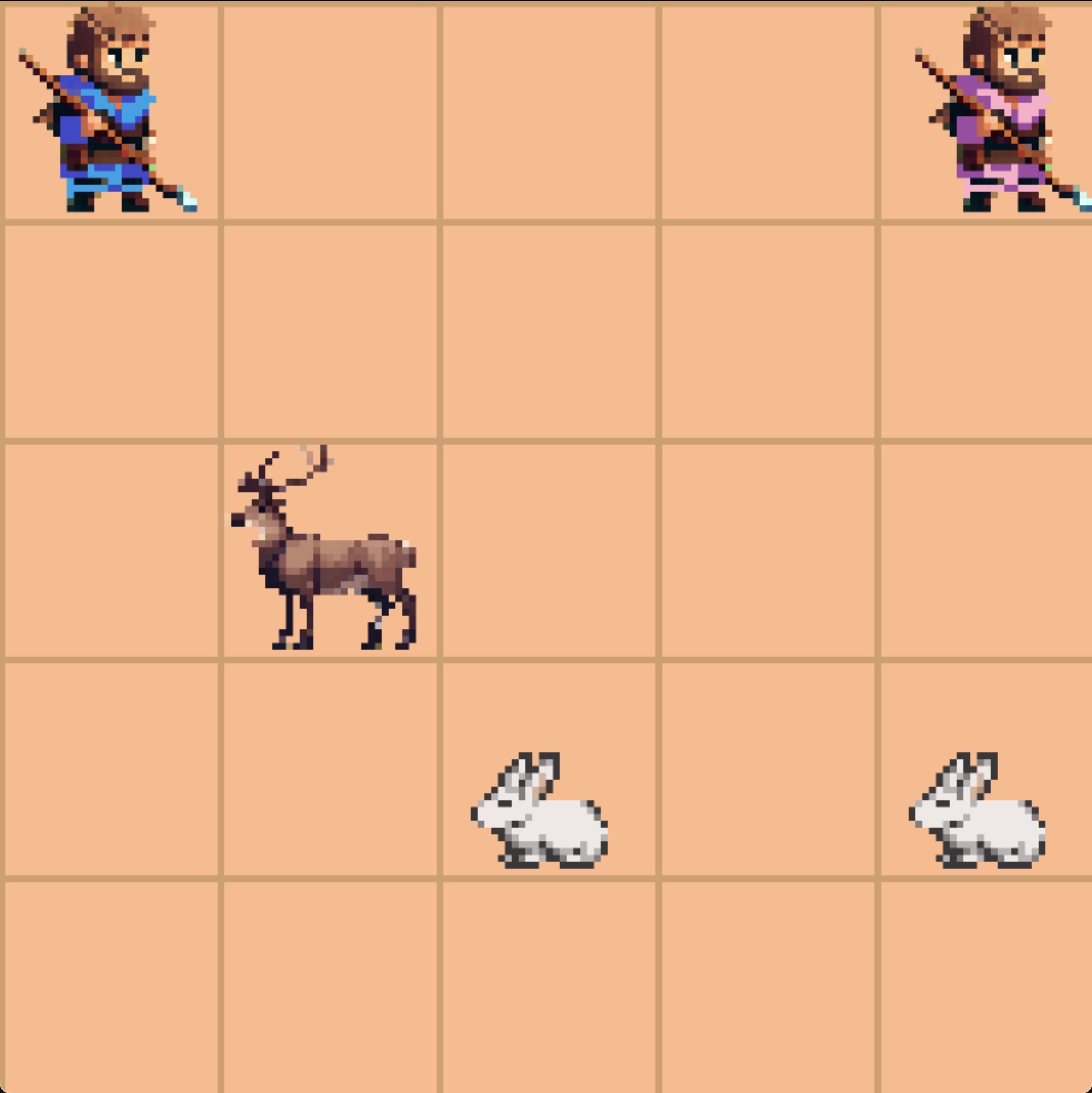}
    \caption{Example grid-world configuration showing the human agent (blue, B), machine agent (purple, P), stag (S), and hares (H).}
    \label{fig:grid-world}
\end{figure}

We implement our simulation in Python using a custom PettingZoo~\cite{terry2021pettingzoo} environment. As illustrated in Figure~\ref{fig:grid-world}, the game is played on a $5 \times 5$ grid containing:
\begin{itemize}
    \item Two hunters (blue and purple agents)
    \item One stag (high-value target requiring cooperation)
    \item Two hares (low-value individual targets)
\end{itemize}

Agents observe the environment and must choose between:
\begin{itemize}
    \item Target \textbf{Stag} ($\rightarrow S$).
    \item Target \textbf{Hare} ($\rightarrow H$).
\end{itemize}

The reward function follows classic stag hunt game theory \citep{skyrms2003stag} payoff matrix:
\begin{equation}
    R(a_1, a_2) = 
    \begin{cases}
        (5, 5) & \text{if } a_1 = \text{Stag},\ a_2 = \text{Stag} \\
        (1, 0) & \text{if } a_1 = \text{Stag},\ a_2 = \text{Hare} \\
        (0, 1) & \text{if } a_1 = \text{Hare},\ a_2 = \text{Stag} \\
        (1, 1) & \text{if } a_1 = \text{Hare},\ a_2 = \text{Hare}
    \end{cases}
\end{equation}

The stag hunt has been increasingly adopted in reinforcement learning as an environment for studying cooperation, since it requires agents to trade off between low but safe individual rewards and higher payoffs that depend on coordination \citep{fang2002,peysakhovich2017,tang2021,Acharya2024}.

The first challenge in working with LLM agents was to describe the observation space and state of the game to the LLM agent. We considered a multi-modal approach where snapshots of the grid-world were used as input to the generative model, or providing the x, y coordinates of the entities in the grid-world, or using the relative distances between the hunters, stag, and hares. We hypothesized that the latter approach would be more interpretable for LLMs and could simplify the learning process. Further details regarding the relative distance calculation and prompt engineering are provided in the following subsections.

\subsection{Large Language Models}
We use the following LLMs: Llama 3.1 8B \cite{llama2024dubey}, Mixtral 8x22B \cite{mixtral2024jiang}, and Llama 3.1 70B \cite{llama2024dubey}. These open-source models were chosen to ensure transparency and reproducibility, and to represent a spectrum of models, ranging from efficient, smaller models to large ones developed by different research groups. Table \ref{tab:model_parameters} shows the parameters selected to balance deterministic outputs (temperature = 0) with diverse and high-probability options (Top-P = 0.9 or 0.95) while capping the response length to 1024 tokens to ensure concise, controlled outputs from various model sizes. All prompting and interactions with the LLM agents were performed using the HuggingFace library \cite{huggingface2024hubcli} with Python scripts, allowing for a seamless integration between our experiment environment and the models.

\begin{table}[htbp]
    \centering
    \caption{LLM Parameters for Policy-Agnostic Proxy Evaluation.}
    \label{tab:model_parameters}
    \begin{tabular}{@{}llccc@{}}
        \toprule
        \textbf{Model} & \textbf{Size} & \textbf{Temp.} & \textbf{Top-P} & \textbf{Tokens} \\
        \midrule
        Llama 3.1 & 70B & 0 & 0.9 & 1024 \\
        Mixtral & 8x22B & 0 & 0.9 & 1024 \\ 
        Llama 3.1 & 8B & 0 & 0.95 & 1024 \\
        \bottomrule
    \end{tabular}
\end{table}

\subsection{Human Benchmark Data}
\label{subsec:benchmark_data}
We evaluate LLM-generated decisions against a human benchmark derived from a study by \citet{Baber2024}, in which 30 participants with minimal exposure to game theory made decisions across 15 grid configurations using a layout similar to the one described in Section~\ref{subsec:grid_world}. In addition to the 15 × 30 human participant decisions, the dataset also includes choices made by judges with expertise in game theory. Our goal is to use the same data to compare human decisions with those generated by an LLM.

Our methodology repurposes the original grid configurations in our environment to test the LLMs agents. As mentioned in Section \ref{subsec:grid_world}, we represent the state of the environment by summarising the relative distances between objects in the grid-world.  We do this to minimise the need for calculation or image interpretation in the LLM; as our concern is with defining the strategy rather than exploring ability to analyse patterns in an image, we assume this is a reasonable step to take. Relative distances provide a more direct encoding of strategic relationships between agents and targets than grid coordinates, which require additional inference of proximity and risk bottlenecking smaller LLMs with spatial interpretation rather than decision-making. From the 15 grid-world scenarios used in \citet{Baber2024}, the x, y coordinates of each object were extracted, and four key features were calculated: the Manhattan distance between pairs of objects: (1) the human player (blue hunter) and the hare closest to it (B-H), (2) the human player (blue hunter) and the stag (B-S), (3) the computer player (purple hunter) and the hare closest to it (P-H), and (4) the computer player (purple hunter) and the stag (P-S). This enables direct comparison between synthetic (LLM) and organic (human) decision-making in policy-agnostic collaboration scenarios.

\section{Experiment 1: Can LLMs replicate expert judges' decisions with full observability of the environment?}
\label{sec:alignment_with_judges}
We evaluate LLM performance against the expert judges’ decisions using 15 grid configurations from \citet{Baber2024} as detailed in Section \ref{subsec:benchmark_data}. To ensure consistency with the human benchmark, we reused the same configurations and decision criteria.

\subsection{Prompt Design}
\label{sec:prompt_design}
Prompts are programmatically constructed as follows:
\begin{enumerate}
    \item \textbf{Game State Extraction}: Retrieve agent and target coordinates from the grid-world environment.
    \item \textbf{Feature Calculation}: Compute Manhattan distances (B-H, B-S, P-H, P-S) and directions.
    \item \textbf{Template Filling}: Inject dynamic features into a pre-defined prompt template.
\end{enumerate}

\textbf{Example Prompt:}
\begin{quote}
\textit{"You are playing a stag hunt game where you earn 5 points for hunting a stag with the second player and 1 point for capturing a hare. You are the Blue player, B, and the other player is purple, P.\\
The distance between you and the nearest hare (B-H) is 2.\\
The distance between you and the stag (B-S) is 5.\\
The distance between the second player and their nearest hare (P-H) is 2.\\
The distance between the second player and the stag (P-S) is 1.\\
Based on these distances, what do you think your target should be? Stag or Hare?\\
Strictly answer in exactly one word."}
\end{quote}

\textbf{Example Output:}
\begin{quote}
\textit{``Stag''}
\end{quote}

In this example, the prompt is designed to present the LLM with a clear depiction of the game state and prompt it to select an optimal target. The four features provided (\textit{B-H}, \textit{B-S}, \textit{P-H}, \textit{P-S}) give the model the relative distances between objects so that its explanation could imply trade-offs between pursuing the stag or the hare, while the reward structure nudges it toward considering the potential payoffs from cooperation versus defection. The same prompts were provided to the Llama 3.1 8B, Mixtral 8x22B, and Llama 3.1 70B models.

\subsection{Evaluation Metrics}
\label{sec:evaluation_metrics}
We compare LLM-generated decisions to those made by expert judges using the following metrics:
\begin{itemize}
    \item \textbf{Precision:} Proportion of correct stag/hare predictions out of all predicted.
    \item \textbf{Recall:} Proportion of actual stag/hare choices correctly predicted.
    \item \textbf{F1-Score:} Harmonic mean of precision and recall, offering a balance between the two.
    \item \textbf{Cohen’s Kappa ($\kappa$):} A statistic that measures inter-rater agreement between model predictions and expert judges, adjusting for the level of agreement expected by chance. Values range from $1$ (perfect agreement) through $0$ (chance-level agreement) to $-1$ (agreement worse than chance).
\end{itemize}

\subsection{Results from Experiment 1}

Larger models, LLaMA 3.1 70B (Avg F1 = 0.80, $\kappa = 0.60$) and Mixtral 8x22B (Avg F1 = 0.79, $\kappa = 0.58$), closely align with expert judgments, substantially outperforming the smaller LLaMA 3.1 8B model (Avg F1 = 0.35), which dominantly outputs \textit{Stag} and generalises poorly (table~\ref{tab:accuracy}). Mixtral demonstrates high precision for \textit{Stag} (0.84) but lower recall (0.68), resulting in an F1-score of 0.75. Macro-average metrics across both top models remain near 0.80, indicating a reliable generalization. In contrast, with human participants (who, from \citet{Baber2024} achieve a Kappa of only 0.07), the LLMs have reasonable kappa scores (table~\ref{tab:kappa_scores}), highlighting the superior consistency of LLaMA 70B and Mixtral.

\begin{table}[htbp]
\centering
\caption{Metrics for LLMs compared with expert judges.}
\label{tab:accuracy}
\begin{tabular}{|l|l|c|c|c|}
\hline
\textbf{Model} & \textbf{Class} & \textbf{Precision} & \textbf{Recall} & \textbf{F1-Score} \\ \hline
\textbf{Llama 3.1 70B} & Hare & 0.84 & 0.78 & 0.81 \\ 
                        & Stag  & 0.76 & 0.82 & 0.79 \\ \cline{2-5}
                        & Macro avg & 0.80 & 0.80 & 0.80 \\
                        & Weighted avg & 0.80 & 0.80 & 0.80 \\ \hline
                                
\textbf{Mixtral 8x22B} & Hare & 0.77 & 0.89 & 0.82 \\
                      & Stag  & 0.84 & 0.68 & 0.75 \\ \cline{2-5}
                      & Macro avg. & 0.80 & 0.78 & 0.79 \\
                      & Weighted avg. & 0.80 & 0.79 & 0.79 \\ \hline

\textbf{Llama 3.1 8B} & Hare & 0.47 & 0.10 & 0.16 \\
                      & Stag  & 0.44 & 0.87 & 0.59 \\ \cline{2-5}
                      & Macro avg. & 0.46 & 0.48 & 0.37 \\
                      & Weighted avg. & 0.46 & 0.45 & 0.35 \\ \hline
\end{tabular}
\end{table}

\begin{table}[htbp]
    \centering
    \caption{Cohen's Kappa Scores for Model Performance.}
    \begin{tabular}{|c|c|}
        \hline
        \textbf{Model} & \textbf{Cohen's Kappa Score} \\
        \hline
        Llama 3.1 70B & 0.599 \\
        \hline
        Mixtral 8x22B & 0.576 \\
        \hline
        Human Baseline & 0.067 \\
        \hline
    \end{tabular}
    \label{tab:kappa_scores}
\end{table}

\subsection{Discussion from Experiment 1}
\label{subsec:discussion_exp1}

These results demonstrate that LLMs, particularly larger models like LLaMA 3.1 70B (with temperature set to 0), achieve expert-aligned decision-making, validating their utility as scalable, policy-agnostic proxies for cooperative tasks.

\begin{equation}
    \text{Alignment: } \overline{\text{F}_1}(a_{\text{LLM}} = a_{\text{expert}} \mid s) \geq 0.75 \tag{Q1}
\end{equation}

The prompts reduce the environment to simplified distance features and payoff structures, which may favour models that exploit these cues deterministically rather than apply reasoning in a more flexible or human-like manner. While these findings demonstrate the feasibility of using large LLMs as expert-aligned agents under full observability, they also underscore the need to test their robustness in settings with partial observability and richer state descriptions. We used relative distances rather than global coordinates to mirror the observation space available to humans in \citet{Baber2024}, where participants made isolated stag vs. hare choices on static grid configurations rather than full action trajectories; extending to coordinate-based or sequential decision-making remains an avenue for future work.

\section{Experiment 2: Can prompts to LLMs induce human-like response variability and risk sensitivity?}
\label{sec:risk_behaviour_simulation}

\begin{figure*}[ht]
    \centering
    \includegraphics[width=\textwidth]{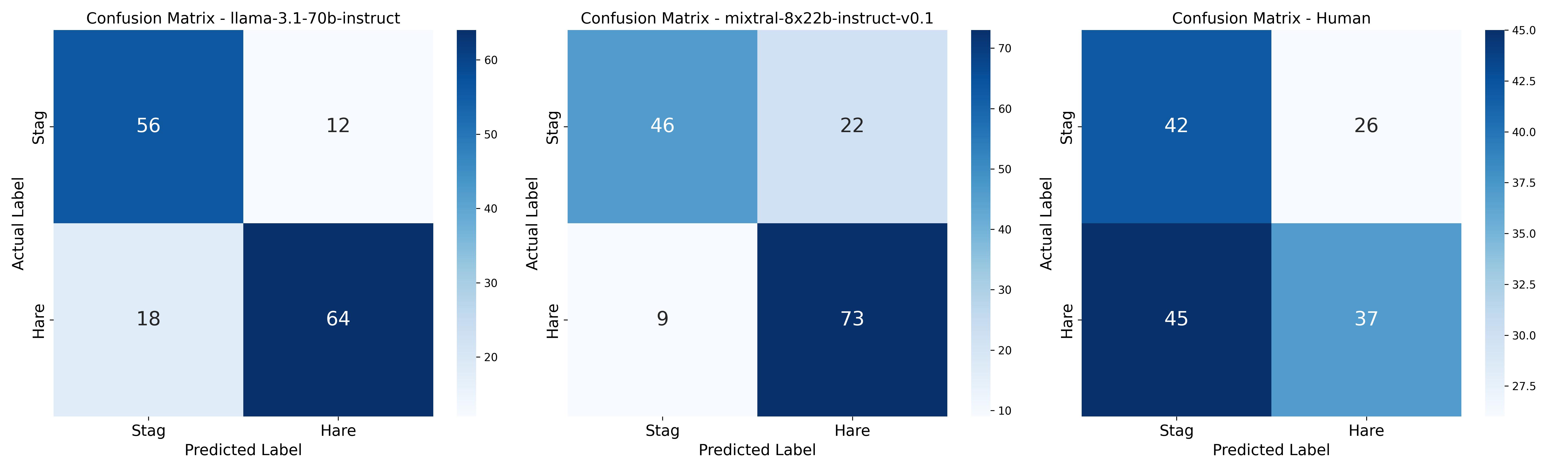}
    \caption{Comparison of Confusion Matrices: Llama 3.1 70B, Mixtral 8x22B, and Human Participants from \citet{Baber2024}.}
    \label{fig:confusion_matrix}
\end{figure*}

To investigate if LLMs can generate decisions with human-like response variability and risk sensitivity, we started by comparing the performance of the Llama 3.1 70B and Mixtral 8x22B models from Section \ref{sec:alignment_with_judges} and the 30 human participants from \citet{Baber2024}. The original study classified the human participants into three categories: "Minimize risk," "Neutral," and "Cooperative," which will be useful for benchmarking \textit{Q2} in our study. The confusion matrix (Figure \ref{fig:confusion_matrix}) highlights the distinct performances of the models. The Llama 3.1 70B model shows balanced accuracy with 56 correct Stag and 64 correct Hare predictions, albeit with 18 false Hare predictions due to a slight over-prediction of Hare (82 Hare vs. 74 Stag). In contrast, the Mixtral 8x22B model exhibits a stronger bias towards Hare, correctly predicting it 73 times but making 22 false Hare predictions and only 46 correct Stag predictions with 9 false Stag predictions. The human baseline has a more balanced distribution with 42 correct Stag and 37 correct Hare predictions, but it also suffers from the highest misclassification rates (45 false Stag and 26 false Hare predictions). Overall, both AI models outperform the human baseline, with Llama 3.1 70B demonstrating the most neural behaviour and balanced accuracy, while Mixtral 8x22B leans more towards predicting Hare possibly indicating an implicit risk averse behaviour.

Building on the approach outlined in Section \ref{sec:alignment_with_judges} and to answer (Q2), we further modified the prompts to simulate different human decision strategies as reflected in preferences for risk and reward. By comparing the decisions made by risk-averse versus risk-seeking LLM agents, we aim to understand how closely their decision-making resembles human behaviour under uncertainty. Additionally, we investigate whether certain configurations consistently lead to more cooperative or defecting behaviour.

\subsection{Modified Prompt Design}
In this version, the structure of the prompt remains similar to that described in Section \ref{sec:prompt_design}, but now includes a risk behaviour modifier specifying risk aversion or risk seeking. The addition of this information is intended to influence the decision-making strategy of the LLM and provides a more nuanced simulation of human behaviour. This change is particularly relevant in scenarios where cooperation (hunting the stag) carries greater uncertainty or risk compared to defecting (capturing the hare).

\textbf{Example Prompt:}
\begin{quote}
\textit{"You are playing a stag hunt game where you earn 5 points for hunting a stag with the second player and 1 point for capturing a hare. You are the Blue player, B, and the other player is purple, P.\\
\textbf{You are risk averse.}\\
The distance between you and the nearest hare (B-H) is 2.\\
The distance between you and the stag (B-S) is 5.\\
The distance between the second player and their nearest hare (P-H) is 2.\\
The distance between the second player and the stag (P-S) is 1.\\
Based on these distances, what do you think your target should be? Stag or Hare?\\
Strictly answer in exactly one word."}
\end{quote}

\textbf{Example Output:}
\begin{quote}
\textit{``Hare''}
\end{quote}

In this example, the added line \textit{"You are risk averse."} informs the LLM to adopt a cautious strategy, likely leading it to prioritise lower-risk options, such as capturing a hare, even though the reward is lower. Conversely, if the LLM were instructed to be \textit{"risk-seeking"} it might prefer the stag, despite the greater uncertainty involved in hunting it.

\subsection{Evaluation Metrics for Stag Hunt Strategy}
\label{sec:evaluation_metrics_risk}

To assess the influence of the risk preferences on the LLM's decisions, we define a \textit{Risk Index} ($\phi_\text{risk}$):

\[
\phi_{\text{risk}} = \frac{N_{\text{Hare}} - N_{\text{Stag}}}{N_{\text{Total}}}, \quad \phi_{\text{risk}} \in [-1, 1]
\]

where $N_{\text{Hare}}$ and $N_{\text{Stag}}$ are counts of decisions to defect (capture hare) or cooperate (hunt stag), respectively, and $N_{\text{Total}} = 15$ is the total number of decisions. Negative values of $\phi_{\text{risk}}$ indicate a tendency toward cooperation (risk-seeking stag hunts), positive values indicate a tendency toward defection (risk-averse hare hunts), and values close to zero suggest balanced behaviour. We set thresholds of $\pm 0.2$ to capture clear deviations from neutrality while leaving a central zone to represent near-equal stag and hare selections, providing a simple but interpretable categorization of strategic bias.

\subsection{Results from Experiment 2}
\begin{figure*}[ht]
    \centering
    \includegraphics[width=0.8\textwidth]{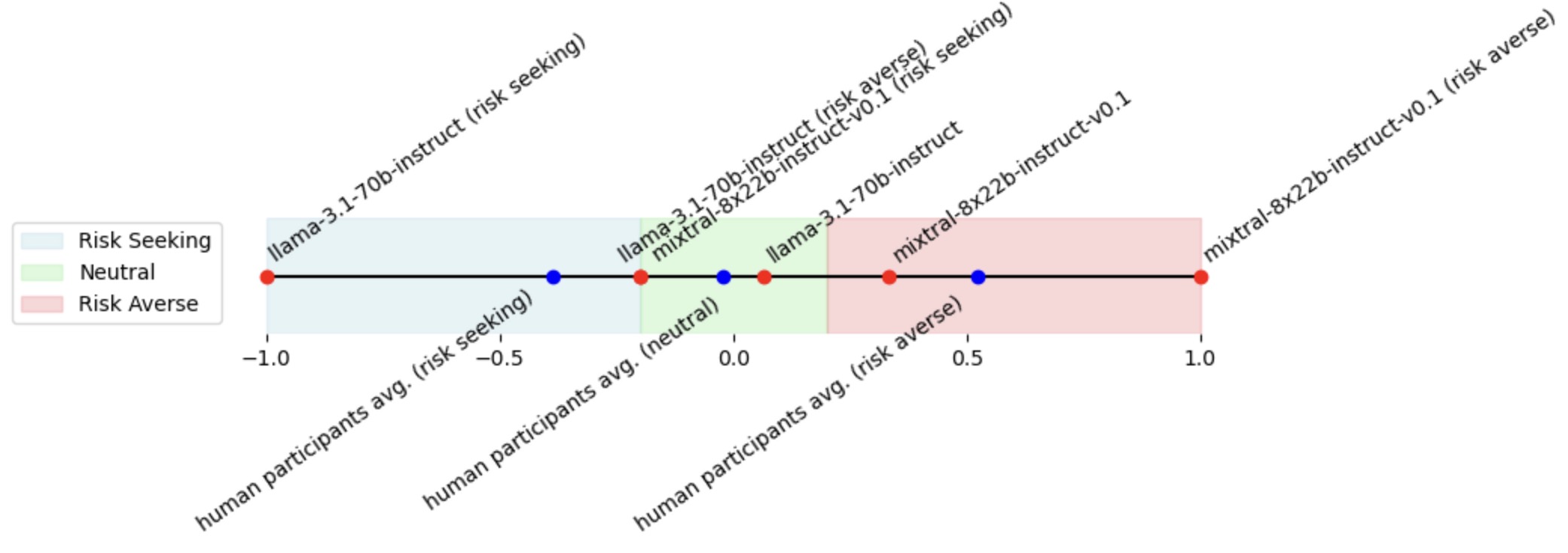}
    \caption{Human and model risk behaviours ($\phi_\text{risk}$) across risk-seeking, neutral, and risk averse ranges (-1 to 1), with positions reflecting varying decision-making tendencies.}
    \label{fig:risk_behaviour}
\end{figure*}

Figure \ref{fig:risk_behaviour} shows the distribution of model behaviours ($\phi_\text{risk}$) along a range from -1 to 1, highlighting distinct decision-making tendencies. The Risk Seeking range (-1 to -0.2) includes models with risk-seeking behaviour, while the Risk Averse range (0.2 to 1) captures more cautious models. The Neutral range (-0.2 to 0.2) represents models with balanced decision-making tendencies. Individual models are annotated with their respective positions, and risk preferences are indicated in parentheses.

\subsection{Discussion from Experiment 2}
The results indicate that both models can simulate risk-averse and risk-seeking behaviours with minor prompt modifications. Llama 3.1 70B defaults toward neutral or cooperative (risk-seeking) behaviour, while Mixtral 8x22B exhibits a more risk-averse baseline but adapts effectively when guided by prompt modifiers. This flexibility suggests that LLMs can be steered to reflect different risk profiles relevant to team coordination, rather than serving as generic human analogues. 

Our use of lightweight prompt engineering shows that context alone can shift model behaviour, highlighting both the adaptability and sensitivity of LLMs to framing. However, this also underscores a limitation: behaviours observed here may not generalise to richer environments or larger groups of agents without redesigning the observation space and prompts. We view such redesign as a lightweight extension rather than a fundamental barrier, aligning with our broader focus on LLMs as proxies in heterogeneous teams.

\section{Experiment 3: Can LLM agents simulate human-like decision-making and generate coherent multi-step action sequences in a multi-agent team?}
\label{sec:LLM_for_imitation}
We investigate whether LLM agents, tuned with the prompts discussed earlier, can simulate specific decision-making behaviours and perform similarly to humans. Specifically, we explore how LLM agents can function as policy-agnostic agents-in-the-loop. To address this, we evaluate the performance of the following models: Mixtral 8x22B and Llama 3.1 70B, in their neutral, risk-seeking, and risk-averse variants, and benchmark their action sequences against those taken by humans in identical states.

\subsection{Collecting Data from Human Participants}
\label{sec:collecting_human_data}
Using the custom PettingZoo environment of our design detailed in Section \ref{subsec:grid_world}, we simulated different game scenarios. For all scenarios, the Blue Hunter started in the top left cell of a 5 x 5 grid, and the Purple Hunter started in the top right cell.  In each scenario, two hares and one stag re-spawned at random. The Blue Hunter was controlled by human participants (using W, A, S, D, X keys to move up, move left, move down, move right and stay). The Purple Hunter followed a predefined script to move towards either hare or stag (on a 70:30 split) and moved immediately after the Blue Hunter. This gave the human participants the impression that the Purple Hunter was responding to their actions. Participants were not informed of the Purple Hunter preference for hare or stag prior to the experiment. When each hunter had arrived at a target (hare or stag) the game reset, with Purple and Blue Hunters in their starting positions and hares and stag spawning in new locations.

We recruited 10 participants for this exercise. Our aim was to collect sufficient data to explore variation in paths chosen to the targets. The data collection exercise was covered by University of Birmingham Ethics Approval Board (ERN 22 1145). Participants each played 9 scenarios of the game, with a Tobii Pro Fusion Eye-tracker recording gaze data and pupil dilation, running at 60Hz, and the position of objects on the screen logged for each scenario. In this paper, we are only considering the position of objects.


\subsection{Collecting Data from LLM Agents}
For the evaluation of LLM agents, we tested them in the same custom environment as described in Section \ref{sec:collecting_human_data}. In each scenario, the Blue Hunter was controlled by the LLM, which was queried to generate the next action based on the current state. The Purple Hunter continued to follow a predefined script, as described in the human participant setup. The agents interacted with the environment, and we collected the resulting state-action pairs for each episode, capturing the decision-making trajectory of the LLM agents.

We tested multiple variants of LLM models (Mixtral 8x22B and Llama 3.1 70B) under different behavior profiles (neutral, risk-seeking, and risk-averse). Data collection included the sequence of state-action pairs over multiple trials to assess the decision-making patterns of the LLM agents and their ability to simulate human-like behaviour in the environment, as outlined in Section \ref{sec:collecting_human_data}. In order to define the observation space and trajectories in a dynamic game, the LLM needed to be queried at each state of the environment to decide the next action \( a_t \) from a predefined action space \texttt{A}. The environment transitions based on the LLM's decisions, creating real-time action-state pairs that guide the trajectory of the agent. The LLM directly generates actions in response to the current state \( \texttt{S}_t \), and these actions dictate how the environment evolves. This can be viewed as a form of in-the-loop decision-making, where the LLM acts as the agent's decision policy throughout the episode. The action-state is record to form trajectories that can be fed into imitation learning algorithms to train agents to mimic human-like or expert behaviour.

In this setup, the LLM is responsible for generating actions in the following iterative process on a custom stag hunt environment:

\begin{enumerate}
    \item \textbf{State Observation}: At each time step \( t \), the current state of the environment \( \texttt{S}_t \) is captured. This includes key features like the positions of the agents, stag, and hares (i.e., variations of \textit{B-H}, \textit{B-S}, \textit{P-H}, and \textit{P-S} with direction) as described in Section \ref{sec:prompt_design}. These features encapsulate the environment’s dynamics, providing the necessary information for the LLM to make a decision.

    \item \textbf{LLM Action Generation}: Given the current state \( \texttt{S}_t \), the LLM is queried to choose an action \( a_t \in \texttt{A} \). The action space \( \texttt{A} \) typically consists of possible moves the agent can make, such as:
    \begin{quote}
        \textit{'UP' - move up, 'LEFT' - move left, 'DOWN' - move down, 'RIGHT' - move right, 'STAY' - stay in place.}
    \end{quote}
    Based on the prompt, which conveys the current state, the LLM selects one of these actions to execute.

    \item \textbf{Action Execution and Environment Update}: After the LLM selects an action \( a_t \), it is executed in the environment. The environment then transitions to a new state \( \texttt{S}_{t+1} \) based on the action taken. The reward structure (e.g., the agent receives higher rewards for hunting the stag together and lower rewards for capturing a hare) further drives the dynamics of the environment, informing the LLM's future decisions.

    \item \textbf{Next State and Trajectory Formation}: The updated state \( \texttt{S}_{t+1} \) is then fed back into the LLM, which continues to make decisions at each subsequent time step. This process repeats iteratively until a terminal state is reached (e.g., after a predefined number of steps or when the game ends). The full sequence of state-action pairs, \( \{(\texttt{S}_t, a_t), (\texttt{S}_{t+1}, a_{t+1}), \dots \} \), forms a trajectory that represents the LLM's decision-making process across the entire episode.
\end{enumerate}

\subsubsection{Prompt Design}
\label{sec:prompt_design_trajectory}
\textbf{Example Prompt:}
\begin{quote}
\textit{"You are playing a stag hunt game where you earn 5 points for hunting a stag with the second player and 1 point for capturing a hare. You are playing as the Blue player, B, and the other player is Purple, P.\\
You are risk seeking.\\
You can choose from the following actions:\\
LEFT, RIGHT, DOWN, UP, STAY\\
Here is the current observation:\\
The hare nearest to you is 2 cells to the right and 2 cells down.\\
The stag 4 cells to the right and 1 cell down.\\
For the second player, the nearest hare 1 cell to the left and 2 cells down.\\
For the second player, the stag is 1 cell down.\\
What action should you take? (LEFT, RIGHT, DOWN, UP, STAY)\\
Strictly answer in exactly one word."}
\end{quote}

\textbf{Example Output:}
\begin{quote}
\textit{``RIGHT''}
\end{quote}

By continuously querying the LLM in response to the dynamically changing environment, we generate decision trajectories that exhibit consistent decision-making behaviors, such as risk-averse or risk-seeking tendencies, across multiple iterations of the stag hunt game. These trajectories can then be compared with data generated by human participants.

\subsection{Results from Experiment 3}

\begin{figure}[htbp]
    \centering
    \includegraphics[width=0.99\linewidth]{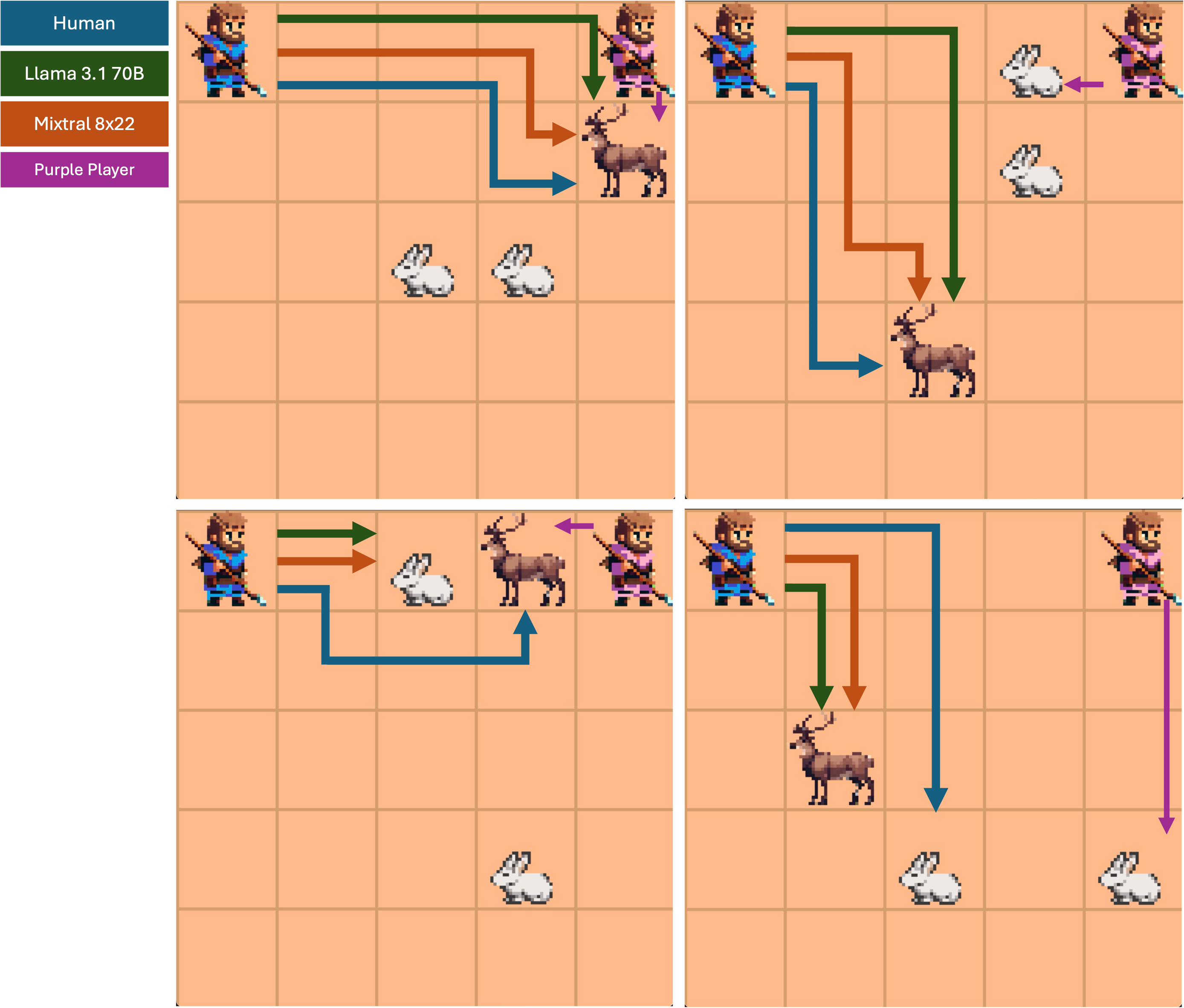}
    \caption{Movement trajectories of Human (Blue), Llama 3.1 70B (Green), Mixtral 8x22 (Red), and Purple Hunter (Purple) in a 5x5 dynamic stag hunt environment. The Blue Hunter is controlled by human and LLM agents, while the Purple Hunter follows a scripted path. The LLM models demonstrate varying degrees of imitation of human decision-making patterns, with Llama 3.1 showing strong alignment in risk-seeking behaviours.}
    \label{fig:decision_paths}
\end{figure}

Figure \ref{fig:decision_paths} illustrates the decision trajectories generated by the LLM agents (Mixtral 8x22B and Llama 3.1 70B) compared to human player decisions, based on the described dynamic stag hunt environment. The visualised data consists of movement trajectories for each agent in a 5x5 grid, where the Blue Hunter is controlled by a human or LLM participant and the Purple Hunter is modelled based on pre-defined scripts as detailed in Section \ref{sec:collecting_human_data}.

The paths taken by the human players and the LLM agents are compared in terms of their movement decisions towards either the hares or the stag. The main focus was to observe if the LLM agents could replicate the movement trajectories of human players, in terms of plausible but varied routes to a target.

\subsubsection{Movement Trajectory Patterns}
\begin{itemize}
    \item \textbf{Human Trajectory (Blue)}: In the depicted scenarios, the human-controlled Blue Hunter exhibits a preference towards the stag. The trajectory shows that human players tend to navigate towards higher-reward targets, demonstrating risk-seeking behaviour by prioritizing the stag over the hares. This is most evident in the bottom-left scenario where the human player moves around the hare to go towards the stag.
  
    \item \textbf{Llama 3.1 70B (Green)}: This model follows a relatively similar trajectory to the human player. In several scenarios (e.g., top-right and top-left), the model chooses to pursue the stag, mimicking the risk-seeking strategy adopted by human participants.
  
    \item \textbf{Mixtral 8x22 (Orange)}: The Mixtral model displays pretty similar decision-making behaviour to the Llama model though with slightly different paths.
  
    \item \textbf{Purple Hunter (Purple)}: The Purple Hunter, following a scripted strategy with a preference for the stag, consistently moves towards the stag in each scenario. This is a predefined path and serves as a baseline for comparison.
\end{itemize}

\subsection{Discussion from Experiment 3}
Despite occasional deviations from the specific human participant chosen for comparison, the in-the-loop actions generated by both LLMs (Llama 3.1 70B and Mixtral 8x22) appear to exhibit human-like decision making. Each trajectory demonstrates clear intent and goal-oriented behaviour, reflecting decisions that are similar to those a human might make under similar circumstances. 

In certain cases, such as the top-right and top-left scenarios, the LLM-generated trajectories closely resemble those of the human participants, with very similar paths taken. While there are minor differences, the overall number of movements and the final outcomes are remarkably similar, which is precisely the kind of behaviour desired when generating data for training imitation models. Moreover, this slight variability observed in the LLM-generated trajectories mirrors the natural variability found in human actions. 

\section{Conclusion}

This paper evaluated whether large language models (LLMs) can serve as effective human proxies and policy-agnostic teammates in heterogeneous multi-agent reinforcement learning (MARL). Through three targeted experiments in a grid-world capture game inspired by the stag hunt game, we showed:

\begin{enumerate}
    \item \textbf{Alignment with Expert Decisions} In Experiment 1, LLMs—most notably Llama 3.1 70B—matched expert judge labels with over 80\% F1-Score under full observability. This confirms that, when provided with structured state features, LLMs can replicate high-level strategic judgments and function as credible stand-ins for human experts.
    \item \textbf{Induced Behavioural Variability} Experiment 2 demonstrated that lightweight prompt modifications can systematically bias LLMs toward risk-averse or risk-seeking choices. While larger models exhibited a ceiling effect in the constrained $5 \times 5$ grid, this validates our approach for eliciting diverse decision profiles. Varying grid size, object descriptions, or prompt phrasing offers a clear path to mitigate ceiling effects in future work.
    \item \textbf{Human-Proxy Decision Trajectories (Q3).} In Experiment 3, trajectories generated by LLM agents mirrored human strategies, especially under cooperative prompts. These were not identical to human paths but were goal-consistent, highlighting the value of LLMs as policy-agnostic teammates and human proxies in heterogeneous-agent settings.
\end{enumerate}

Collectively, these findings establish LLMs as scalable, customizable proxies for human decision-making in multi-agent settings. Unlike prior work that trains specialised models to simulate humans, our approach is \textit{policy-agnostic}: LLMs act directly from textual prompts rather than pre-trained policies. Prompts are environment-specific but model-agnostic, i.e., only simple features, such as relative distances, were supplied.  This makes design low-effort and amenable to automation. Importantly, LLMs responded to structured observation prompts, not interpretations of agent behaviours, ensuring comparability with human decision labels.

\subsubsection*{Limitations and Future Work}
Our analysis focused on a single grid-world task; generalisation to more complex or continuous domains remains to be demonstrated. The ceiling effect suggests that environment design and prompt granularity warrant further study. Future work will:
\begin{itemize}
    \item Extend the approach to settings with two or more LLM-controlled agents whose prompts update dynamically with the grid-world state.
    \item Integrate LLM-agents into RL training pipelines following \citet{Acharya2024}, replacing pre-defined MDP agents with LLM-based proxies.
    \item Compare LLM agent behaviour directly with human participants.
    \item Incorporate LLM agents alongside RL-trained agents in heterogeneous human-AI teams.
\end{itemize}

Ultimately, this line of work aims to enable seamless human-AI collaboration in open-ended settings by endowing agents with the flexibility and nuance of human decision-making.



\begin{ack}
The work reported in this paper was partly supported by a cooperative agreement award (W911NF-22-2-0161) from the DEVCOM Army Research Laboratory to the Alan Turing Institute and University of Birmingham.
\end{ack}



\bibliography{references}

\end{document}